\pdfoutput=1

\documentclass[11pt]{article}

\usepackage[]{EMNLP2022}

\usepackage{times}
\usepackage{latexsym}

\usepackage[T1]{fontenc}

\usepackage[utf8]{inputenc}

\usepackage{microtype}

\usepackage{inconsolata}

\usepackage{url}
\usepackage{textcomp}
\usepackage{booktabs} 
\usepackage[flushleft]{threeparttable}
\usepackage{multirow}
\usepackage{graphicx}
\usepackage{enumitem}
\usepackage{array}
\usepackage{amssymb}
\usepackage{xspace}
\usepackage{mathtools}
\usepackage{xcolor}
\usepackage{pgfplots}
\usepackage{tabularx}
\usepackage{filecontents}
\usepackage{tikz}
\usepackage{subcaption}
\usepackage{color}
\usepackage{todonotes}
\usepackage{caption}

\usepackage{amssymb,amsfonts,amsmath,amsthm,amscd,dsfont,pifont,mathrsfs}
\usepackage{graphicx,float,epsfig,color,url}
\usepackage{algorithm,algcompatible,algpseudocode,bbm,mathtools}
\usepackage{enumitem}
\usepackage{graphicx}
\usepackage{url}
\usepackage{subcaption}
\usepackage{multirow}

\newcommand{\lcls}{\ell^\text{CLS}_\text{\textcolor{blue}{\text{seg}}}}
\newcommand{\lgenseg}{\ell^\text{GEN}_\text{\textcolor{blue}{\text{seg}}}}
\newcommand{\lgenlabel}{\ell^\text{GEN}_\text{\textcolor{red}{\text{label}}}}
\newcommand{\lgenseglabel}{\ell^\text{GEN}_\text{\textcolor{blue}{\text{seg}} + \textcolor{red}{\text{label}}}}

\title{Structured Summarization: Unified Text Segmentation and Segment Labeling as a Generation Task}

  \author{Hakan Inan, Rashi Rungta, Yashar Mehdad \\ \\
 Meta AI 
 \\
 {\small{\tt \{inan,rashirungta,mehdad\}@meta.com}}
 }

\begin{document}
\maketitle
\begin{abstract}
Text segmentation aims to divide text into contiguous, semantically coherent segments, while segment labeling deals with producing labels for each segment. Past work has shown success in tackling segmentation and labeling for documents and conversations. This has been possible with a combination of task-specific pipelines, supervised and unsupervised learning objectives. In this work, we propose a single encoder-decoder neural network that can handle long documents and conversations, trained simultaneously for both segmentation and segment labeling using only standard supervision. We successfully show a way to solve the combined task as a pure generation task, which we refer to as \emph{structured summarization}. We apply the same technique to both document and conversational data, and we show state of the art performance across datasets for both segmentation and labeling, under both high- and low-resource settings. Our results establish a strong case for considering text segmentation and segment labeling as a whole, and moving towards general-purpose techniques that don't depend on domain expertise or task-specific components.
\end{abstract}

\section{Introduction}

\begin{figure}[ht!]
\centering
\includegraphics[width=1.3\linewidth]{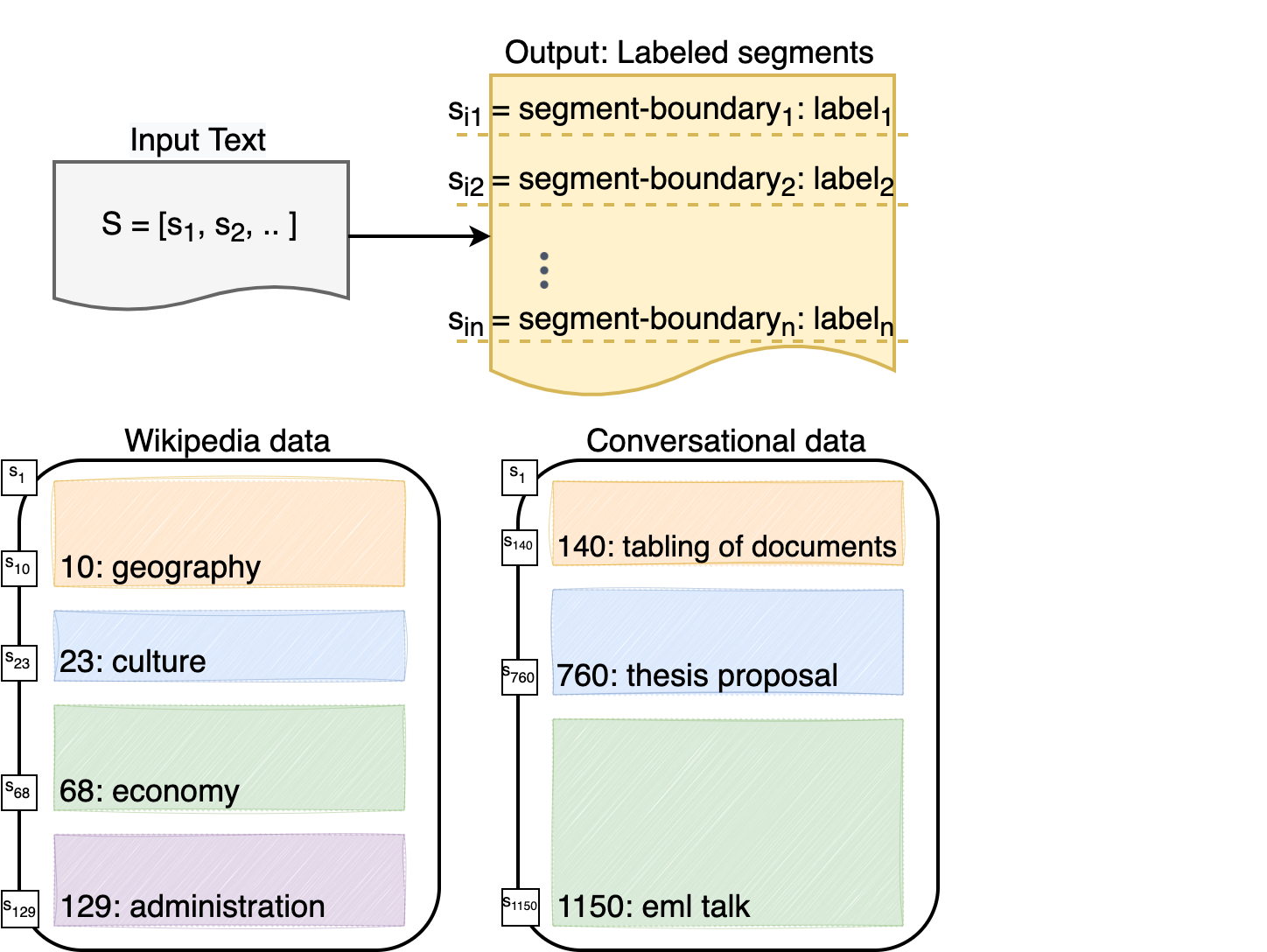}
\caption{Text segmentation and segment labeling. The input text is depicted as a list of sentences \emph{S}. Given $n$ segments, the task output comprises of \emph{ [($s_{i_1}$: $\text{label}_1$), $\ldots$, ($s_{i_n}$: $\text{label}_n$)]}. Shown are labeled segments from two examples of input text, along with the sentence positions that mark the segment boundaries.}
\label{fig:task}
\end{figure}

Text segmentation is the task of organizing text (documents or conversations) into contiguous segments that are individually coherent. This problem can be regarded as a standalone task; however, in practical applications, one is typically interested in labeling each segment with a description. The latter problem is akin to the familiar task of text summarization, applied to each segment individually. The two tasks together can then power more practical applications: dividing a long news piece into sections accompanied by section-level headlines, turning an informational document into topic segments with titles for each topic, generating an outline for conversation transcripts, etc.

The segmentation problem was initially tackled as an unsupervised problem, by computing \emph{coherence} scores between consecutive pairs of sentences in text, and determining segment boundaries according to certain procedures \citep{hearst1994multi}. More recently, text segmentation has been recast as a purely supervised problem, where a neural network can learn segment boundaries from direct dataset supervision \citep{koshorek2018text}. These models perform well, especially in the presence of a large amount of training data.
Despite the success of the more general-purpose supervised approaches, unsupervised objectives and task-specific procedures are still competitive in low-resource settings, such as for conversation segmentation \citep{xing2021improving, solbiati2021unsupervised}. Conversations tend to be longer and less information-dense compared to expository text like from Wikipedia, which may provide additional reasoning for modality-specific treatment to segmentation. Currently, there is not a single, general-purpose approach that works across text modality and training data volume.
Further, due to the long nature of input text, existing methods typically adopt specific mechanisms, such as two-level encoding schemes \citep{koshorek2018text, arnold2019sector, xing2020improving, solbiati2021unsupervised}. 

In the context of segmentation, label generation is thought of as a boundary-constrained summarization or classification problem, with freely generated or categorical labels respectively. In either case, the segment labels are produced using segmentation-aware task components, like segment-restricted attention mechanisms \citep{zhang2019outline}. We note that despite the unifying pressure of the practical applications mentioned above, the label generation task has not been well-coupled to the text segmentation literature, and hence there are fewer works where both segmentation and segment labeling are tackled in a unified context. 

In this work, we argue for the unification of the segmentation and segment labeling tasks, and demonstrate that the two can naturally be put in a monolithic and general-purpose framework while mutually benefiting each others' performance. We start by establishing a simple yet strong baseline for segmentation using a single, long-input transformer encoder trained only with segmentation supervision. We then introduce \emph{generative segmentation}, a novel approach to segmentation as a token generation task, and use an encoder-decoder transformer for it. We show a procedure for generative segmentation that matches the performance of the traditional, discriminative, approach. Finally, we put segmentation and segment labeling on equal footing by generating the segment boundary indices as well as labels in the same output token sequence. We show that this approach sets a new state-of-the-art for both segmentation and labeling. Henceforth, we refer to the combined task of generative segmentation + segment label generation as \emph{structured summarization}.

We summarize our contributions as follows:

\vspace{-2mm}
\begin{itemize}[noitemsep]
\item  We introduce a strong baseline for text segmentation that works across text modalities (documents, conversations) and data scale (high- and low-resource) using a monolithic, supervised learning approach.
\item  We introduce generative segmentation, and show that it matches the performance of traditional, discriminative segmentation.
\item  We propose unifying segmentation and labeling into a single \emph{structured summarization} task with a general-purpose text generation scheme, and show it achieves state-of-the-art performance for both segmentation and segment labeling.
\item  We show that one can pretrain models for structured summarization and transfer them to low-resource (including conversational) datasets, removing the need to use multiple objectives or task-specific logic for obtaining state-of-the-art performance.
\end{itemize}

\begin{figure*}[ht]
\centering
\includegraphics[width=01\textwidth]{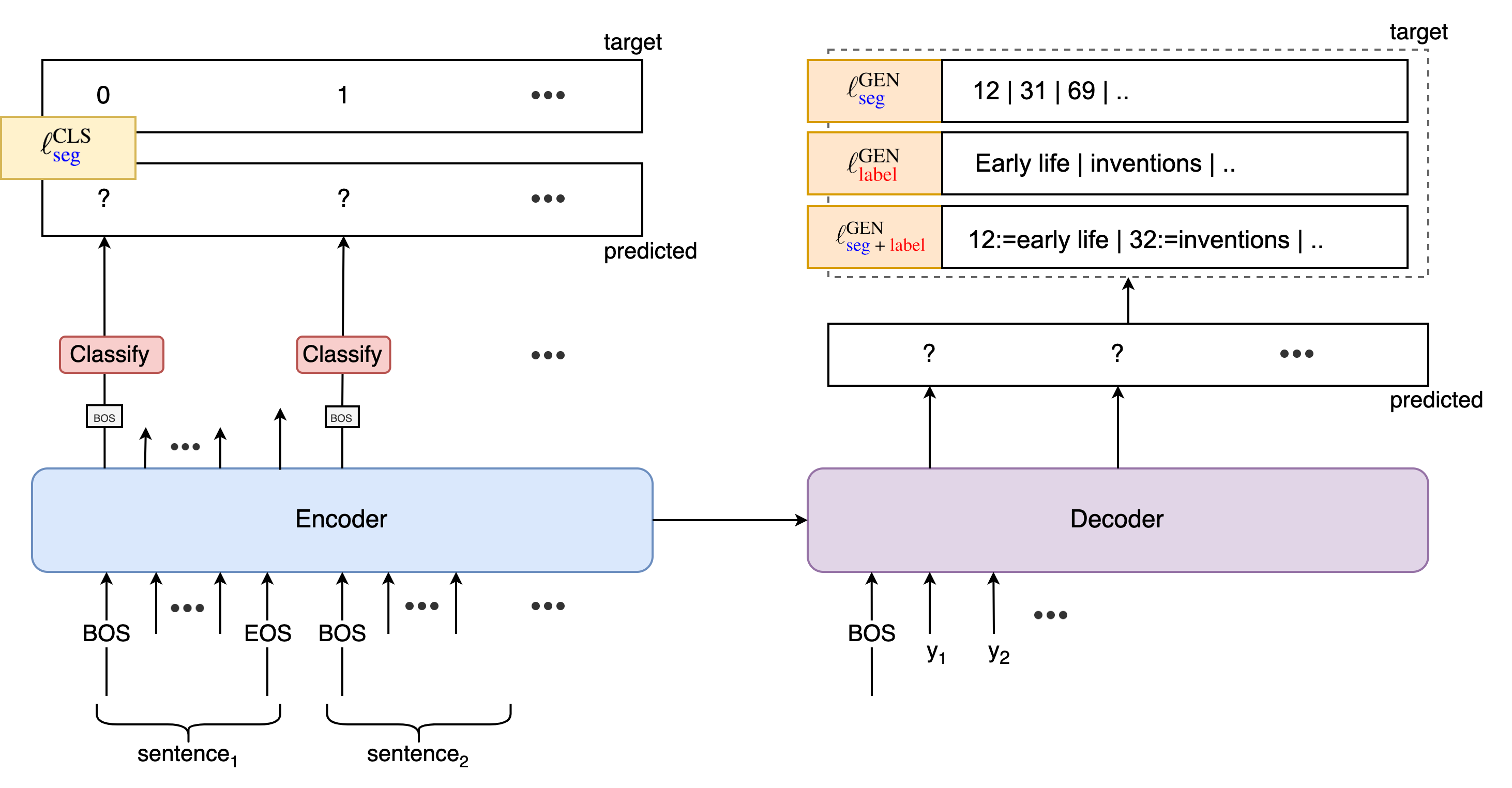}
\caption{Our proposed setup for segmentation and segment labeling. In structured summarization, we use $\lgenseglabel$.}
\label{fig:model}
\end{figure*}

\section{Background: Text Segmentation and Segment Labeling}
\label{sec:background}
In text segmentation, one is given a list of sentences $S=[s_1, s_2, \ldots]$ with length $|S|$. Although several formulations exist, in a commonly accepted formulation, the task is to identify a subset of sentences as \emph{segment boundaries}, which are understood to be the last sentences in each segment. If there are $n$ target segments, the correct output can be understood as a list of $n$ unique (\& monotone) indices,  each index being less than $|S|$. We note that in this work we focus on \emph{linear segmentation} as opposed to \emph{hierarchical segmentation}, wherein one can divide each segment into further sub-segments.

One can tackle the segmentation task in several ways. In low-resource (or no-resource) settings, one can proceed without supervision, as has been done traditionally \citep{hearst1994multi, xing2021improving, solbiati2021unsupervised}. The most common approach computes a \emph{coherence} metric between each consecutive sentence pair (refs) and uses a post-processing method (e.g. TextTiling, \citet{hearst1994multi}) that determines "deep enough" valleys in the resulting coherence vs sentence index curve to draw segment boundaries at certain sentence indices. For instance, coherence can be calculated as the cosine similarity between sentence embeddings given by a large language model. In high-resource settings, one can use standard supervision: calculate embeddings for each sentence, then train a binary classifier on top of each sentence embedding that determines whether the sentence is the last sentence of a segment. Alternatively, one can implement supervision as sequence tagging, with the sentence classifier determining whether a sentence is the beginning or inside of a segment \citep{barrow2020joint}.

Segment labeling is a task that depends on segmentation, and involves producing a descriptive label for each identified segment. It can be implemented as either a discrimination, or generation task. For domains where labels can be binned into predetermined categories (e.g. particular Wikipedia domains with a well-defined ontology), one can implement it as a discriminative task with categorical labels with a fixed vocabulary. We note that the labeling task is discriminative for most prior work that produced or used publicly available datasets \citep{arnold2019sector, barrow2020joint, lo2021transformer}. Given a label vocabulary of size $N$, one can either directly train an $N$-way classifier on top of sentence embeddings \citep{arnold2019sector}, or pool sentence embeddings within a given segment prior to classification \citep{barrow2020joint}. Unfortunately, for a vast number of domains, segment labeling is not easily amenable to a discriminative treatment. These domains include free-form conversations, meeting transcripts, news articles, and even most Wikipedia content. In this case, generation models may be used. For instance, one can utilize sequence-to-sequence neural networks with some form of "segmentation-aware" attention mechanism \citep{zhang2019outline, liu2021end}.

\section{Structured Summarization}
\label{sec:ss}
We view segmentation and labeling as a single task that we refer to as \emph{structured summarization}, guided by the principle of using simpler and more general-purpose approaches, as well as by the wholistic nature of practical applications that involves carrying out both tasks at once. 

In this section, we first describe our encoder backbone that we use as a strong baseline for discriminative segmentation (Section \ref{sec:encoding}). We then introduce generative segmentation (Section \ref{sec:decoding}), which turns segmentation into a sequence generation task. We then combine generative segmentation with segment label generation (Section \ref{sec:combined_seglabel}). Finally, we introduce a recipe for state-of-the-art performance for structured summarization on low-resource datasets (Section \ref{sec:pretraining}).

\subsection{Discriminative Segmentation with a Single Transformer Encoder}
\label{sec:encoding}
As opposed to the common two-level encoding scheme for encoding text as tokens $\xrightarrow{}$ sentences \citep{koshorek2018text, somasundaran2020two, xing2020improving, xing2021improving}, we encode the whole sequence at once with a long transformer.
Recalling that the original sequence was described as a list of sentences $S=[s_1, s_2, \ldots]$ with length $|S|$ (see Section \ref{sec:background}), we expand each sentence in terms of tokens (e.g. SentencePiece, \citet{kudo2018sentencepiece}):
\begin{equation*}
s_i = [t_{i_1}, t_{i_2}, \ldots], \hspace{5mm} i = 1, 2, \ldots, |S|.
\end{equation*}
Thus, the overall sequence becomes
\begin{equation*}
S = [t_{1_1}, t_{1_2}, \ldots \ldots, t_{|S|_1}, \ldots, t_{|S|_m}],
\end{equation*}
where $m$ is the number of tokens in the last ($|S|^\text{th}$) sentence.
We note that the first token of each sentence is a predetermined BOS token index:
\begin{equation}
\label{eqn:bos}
t_{1_1} = t_{2_1} = \ldots  t_{|S|_1} = \text{idx}_{BOS}.
\end{equation}
We then feed the whole sequence $S$ to a transformer encoder. In order to perform discriminative segmentation, we apply a shared binary softmax classifier on top of each BOS token. For training,  we use the binary sentence labels in the dataset $L_{s_i} \in \{0,1\}$ that indicate whether the $i^\text{th}$ sentence is the last sentence of a segment. We use the standard cross-entropy (BCE) loss, which we call $\lcls$:
\begin{equation*}
\lcls = \sum_{i =1}^{|S|} \text{BCE}(h(f(t_{i_1})), L_{s_i}).
\end{equation*}
Above, $h$ refers to the binary classifier, while $f$ refers to the representation of an input token using the neural network encoder. Refer to Figure \ref{fig:model} for a visual description.

\subsection{Generative Segmentation}
\label{sec:decoding}
A major step towards combining segmentation and labeling tasks is to cast segmentation as a generation problem. To this end, we map target segmentation labels into target sequences. We simply turn segment boundary indices into strings, and combine them in a single target sequence. Particularly, given the sentence positions corresponding to segment boundaries, i.e. the ordered set
$pos_{seg} = \{i \mid L_{s_i}=1\}$. We turn the integer-valued positions to strings, and combine them in a single target sequence with delimiters. For instance, in a text with 3 target segments such that $pos_{seg} = [31, 410, 680]$, the target sequence is the string '31 | 410 | 680'. We then turn this string into a list of tokens $\{y_1, y_2, \ldots, y_m\}$ and use as the target tokens. We use an encoder-decoder transformer with the encoder described as in Section \ref{sec:encoding}, and train it with standard teacher forcing. Assuming that the NN decoder outputs a probability distribution over tokens $\hat{p}_i$ at each each time step $i$, and denoting with $1_{k}$ a one-hot (discrete) probability distribution with the $k^\text{th}$ element set to 1, the generative segmentation loss is
\begin{equation*}
\lgenseg = \sum_{i =1}^{m} \text{CE}(\hat{p}_i, 1_{y_i}).
\end{equation*}
Above, CE refers to the standard cross-entropy loss, and $m$ denotes the sequence length in tokens.

\textit{\textbf{Encoding sentence positions.}} We note that the neural network needs to develop an implicit ability to convey position information from input to its decoder, since it is required to produce sentence position as tokens at decoder output. One may suspect that position embeddings may facilitate such ability. On the other hand, it is common for transformer architectures to utilize relative position embeddings (the architecture we use in our experiments, LongT5, falls into this category), without other explicit mechanisms to encode sentence or token positions. We attempt to alleviate this problem from a minimal effort perspective: At the encoder input for the $i^\text{th}$ sentence,  we use the $i^\text{th}$ vocabulary token embedding in place of a fixed BOS token index. Formally, in contrast to \eqref{eqn:bos}, we set
\begin{equation*}
t_{1_1} = 0, \hspace{2mm}  t_{2_1} = 1, \hspace{2mm} \ldots,  t_{|S|_1} = |S|-1.
\end{equation*}
In doing so, we expose the decoder to unambiguous position information that it can exploit for producing sentence indices at the decoder. Importantly, we achieve it without employing custom schemes, such as dedicated sentence position embeddings. In Section \ref{sec:sentpos} of the Appendix, we show that this very simple approach improves substantially over the naive approach, enabling generative segmentation to close the gap with discriminative segmentation.

\textit{\textbf{Post-processing.}} Finally, we post-process the decoder output into a list of segment boundary sentence positions. We split the output by the expected delimiter ("|"), and turn the resulting list of strings into a list of integers, while enforcing they be between 0 and $|S|-1$. Any erroneous part of the output is omitted from the final list of boundary sentence positions. As shown in Section \ref{sec:sentpos_noise} of the Appendix, a well-trained transformer has an erroneous output only about $0.1$ percent of the time.

\subsection{Combined Generative Segmentation + Label Generation}
\label{sec:combined_seglabel}
Having described generative segmentation, we can now combine the segmentation and label information into a single target sequence. Given $n$ target segments for an example text, we first prepare two target sequences: 
\vspace{-2mm}
\begin{itemize}[noitemsep]
    \item $n$ ordered sentence position strings (0-based) separated by a delimiter ("|"), each position indicating the last sentence in a target segment in text. This is optimized via $\lgenseg$ (see Section \ref{sec:decoding}).
    \item $n$ ordered segment labels separated by a delimiter ("|"). We use the same teacher forcing loss as with generative segmentation. We omit details since label generation is a standard seq2seq task, with no special treatment in our approach. We denote the loss for training segment labeling $\lgenlabel$.
\end{itemize}
\vspace{-2mm}
For our canonical structured summarization setup, we simply interweave the two output sequences, aligning the target positions and the target labels (separated with a new delimiter, ":="). The decoder output is trained against $\lgenseglabel$. See Figure \ref{fig:model} for a visual depiction. 
 
\subsection{High-resource Pretraining Prior to Low-resource Training}
\label{sec:pretraining}
Judging from the dominance of unsupervised or loss-augmented methods in prior work in low-resource segmentation and labeling settings, one can argue that monolithic architectures like a single transformer trained purely with dataset supervision don't do well on low-resource datasets. We shall demonstrate this point quantitatively in our experiments (see Section \ref{sec:expts}). For generality and simplicity, it is nonetheless desired to close the performance gap for the monolithic approach trained only with dataset supervision. To this end, we adopt the modern findings from the field that demonstrated the use of pretraining \citep{devlin2018bert, lewis2019bart, brown2020language}. We pretrain our structured summarization models on a large dataset, and treat low-resource training as fine-tuning. In this work, our pretraining dataset (Wiki-727K) has $\sim$0.5M training examples, and we consider scales on the order $10^3$ or less as low-resource. We show in our low-resource experiments that this simple transfer learning setup is enough to not only close the performance gap, but also to set new state-of-the-art. Of note, we are able to translate performance gains from documents to conversations, the latter modality lacking high-resource datasets.

\section{Experiments}
\label{sec:expts}
In this section, we apply our method (see Section \ref{sec:ss}) to 3 datasets, each with distinct characteristics:

{\bf Wiki-727K} \citep{koshorek2018text}: This dataset contains
727,746 English Wikipedia articles, with each article segmented per its existing sections, with segment labels being the section titles. The dataset is split into train/dev/test as 80/10/10 \%. We use this dataset as a high-resource arena to compare structured summarization to existing segmentation methods. We also utilize this dataset for pretraining prior to low-resource training.

{\bf WikiSection} \citep{arnold2019sector}:  This dataset contains 38k full-text documents from English
and German Wikipedia, annotated with sections and topic labels. Differently from Wiki-727K, WikiSection contains normalized topic labels for each section, which are categorical. For that reason, it is amenable to and used for discriminative segment labeling \citep{arnold2019sector, barrow2020joint}. In our experiments, we use the English portion, which includes two datasets corresponding to two Wikipedia domains: 1) \emph{en\_disease}, containing 3.6K total docs, and 2) \emph{en\_city}, containing 19.5K total docs. Both datasets are split into train/dev/test as 70/20/10 \%. We use this dataset as a mid-to-low-resource arena to compare structured summarization to existing segmentation as well as discriminative labeling methods.

{\bf QMSum} \citep{zhong2021qmsum}: This dataset is a collection of long meeting transcripts from academic, product, and committee domains. Among other useful data, each example contains manually curated topic segments and accompanying topic labels. It contains a total of 232 examples split into train/dev/test roughly as 68/16/16 \%. We use this dataset as a low-resource conversational arena, in which we compare structured summarization to existing segmentation approaches.

Throughout, we use the following segmentation and labeling metrics to judge performance:

{\bf $\text{P}_k$} \citep{beeferman1999statistical}: A widely used segmentation metric, $P_k$ estimates the probability that randomly drawn sentence indices that are $k$ sentences apart fall into non-agreeing segments between the reference and predicted segmentation. Keeping with the common practice, we set $k$ equal to half the mean segment length of the reference.

{\bf Rouge} \citep{lin2004rouge}: This is a standard text similarity metric used for summarization. In order to compute Rouge metrics, we gather all predicted and target segment labels and serialize them with a delimiter between labels. We use Rouge-1 (R-1), Rouge-2 (R-2), and Rouge-L (R-L).

{\bf Label F1}: WikiSection dataset contains categorical segment labels. For this reason, we compute the micro-averaged F1 as a measure of accuracy for our generated labels in alignment with prior work. We first align the generated segments to the target segments to have a many-to-one mapping between the two sets of segments. This is done by looking at the closed segment boundary to the left and the right of the predicted segment, mapping the predicted segment to the target segment with maximum overlap. The label for this target segment is then considered the ground truth label for the predicted segment during the F1 calculation.
\vspace{2mm}

\textit{\textbf{Model.}} For all our experiments, we use Long T5-base with transient global attention \citep{guo2021longt5}. We were not able to set up a reliable workflow with large-sized models due to the limits of our computational setup. On the other hand, our hypotheses are testable using any-sized transformer of sufficient expressibility, and our results are self-contained with fair comparisons including strong baselines of our own with same-sized setups.  We use AdamW \citep{loshchilov2017decoupled} with a learning rate of $0.0005$ for all experiments. We use a batch size of 128, 64, and 16 for Wiki-727K, WikiSection, and QMSum respectively. We use maximum sequence length of $16384$ for Wiki-727K and WikiSection, $32768$ for QMSum.

\begin{table}[t]
\small
\centering
\renewcommand{\arraystretch}{1.4} 
\begin{tabular}{l|c|cccc}
\toprule
\multirow{2}{*}{Model}  & CLS & \multicolumn{4}{c}{GEN} \\ 
&  $\text{P}_k$ $\downarrow$ &  $\text{P}_k$$\downarrow$  &  R-1$\uparrow$ &  R-2$\uparrow$ &  R-L$\uparrow$\\
 \hline 
$\lcls$ (DS) & 15.4 & - & - & - & - \\ 
$\lgenseg$ & - & 15.8 & - & - & - \\
$\lcls + \lgenseg$ & 15.5 & 15.8 & - & - & - \\
$\lgenlabel$ & - & - & 48.7 & 28.2 & 48.2 \\
$\lcls +\lgenlabel$ & 15.4 & - & 48.7 & 28.1 & 48.3 \\
$\lgenseglabel$ (SS) & - & \bf 15.0 & \bf 49.1 & \bf 28.5 & \bf 48.6 \\
$\lcls + \lgenseglabel$ & \bf 15.0 & \bf 15.0 & 48.7 & 28.1 & 48.2 \\ \bottomrule
 
\end{tabular}
\caption{Results for different combinations of segmentation and labeling losses on Wiki-727K. DS refers to Discriminative Segmentation and SS refers to Structured Summarization. Structured summarization has the best performance across segmentation and labeling. Boldface fonts indicate the best results.}
\label{table:ablations}
\end{table}

\subsection{Validating Structured Summarization}
We train a Long T5-Base on the Wiki-727K training set using all viable combinations of available losses ($\lcls$, $\lgenseg$, $\lgenlabel$, and $\lgenseglabel$). Our aim is to show the merits of the combined approach, while understanding the contributions of different components.
The results are shown in Table \ref{table:ablations}. We make the following remarks:
\begin{itemize}[noitemsep]
\vspace{-1mm}
\item Generative segmentation performance ($\lgenseg$, $\text{P}_k=15.8$) is nearly on par with discriminative segmentation ($\lcls$, $\text{P}_k=15.4$), demonstrating the practical feasibility of generative segmentation.
\item Although combining $\lcls$ and $\lgenseg$ doesn't improve segmentation performance, combining $\lgenseg$ and $\lgenlabel$ (aka structured summarization) leads to best segmentation. This shows that the labeling task improves generative segmentation. In contrast, when labeling is combined with discriminative segmentation ($\lcls + \lgenlabel$), the segmentation performance does not improve.
\end{itemize}
Our results provide evidence that unifying segmentation and labeling as a generative task is more favorable compared to carrying them out separately. In our next set of experiments, we validate structured summarization (using $\lgenseglabel$) against current best methods in high, medium, and low-resource settings, including both document and conversation data.
\begin{table}[t!]
\small
\setlength{\tabcolsep}{5pt} 
\renewcommand{\arraystretch}{1.4} 
\centering
\begin{tabular}{l|l|cccc}
\toprule
\multicolumn{2}{l|}{Model}  &  $\text{P}_k$ $\downarrow$  & R-1 $\uparrow$ & R-2 $\uparrow$ & R-L $\uparrow$ \\
\midrule
\multicolumn{2}{l|}{TextSeg}  &  22.13  & - & - & - \\
\multicolumn{2}{l|}{TLT-TS} & 19.41  & - & - & -\\
\multicolumn{2}{l|}{CATS} & 15.95  & - & - & - \\
\midrule
 
\multicolumn{2}{l|}{DS ($\lcls$)} &  15.4 & - & - & -  \\
\multicolumn{2}{l|}{SS ($\lgenseglabel$)}  & \bf 15.0 &  \bf 49.1 & \bf 28.5 & \bf 48.6 \\
\bottomrule
\end{tabular}
\caption{Wiki-727K Test Set Segmentation Results. TextSeg: \citep{koshorek2018text}, TLT-TS: \citep{somasundaran2020two}, CATS: \citep{somasundaran2020two}. DS refers to Discriminative Segmentation and SS refers to Structured Summarization. 
}
\label{table:wiki}
\end{table}
\subsection{High Resource, Document Setting}\label{high-res}
We first set out to compare structured summarization to existing text segmentation approaches in the high-resource setting. We also aim to establish single transformer-based discriminative segmentation over the standard discriminative segmentation approach, namely using a two-level (token-level$\xrightarrow{}$ sentence-level) encoding. Although we are not aware of a published baseline for segment label generation for Wiki-727K, we also report \emph{Rouge} metrics for our models where applicable.

Table \ref{table:wiki} shows the results. We first note that our single transformer encoder outperforms the best previous model, which uses a two-level transformer augmented with a coherence loss. We also find that structured summarization outperforms discriminative segmentation, setting a new state-of-the-art on Wiki-727K. Overall, these results favor general-purpose methods trained with only dataset supervision.

\begin{table*}[ht!]
\small
\renewcommand{\arraystretch}{1.3} 
\centering
    \begin{tabular}{l|ccccc|ccccc}
    \toprule
    \multirow{2}{*}{Model}  &  \multicolumn{5}{c|}{en\_city} & \multicolumn{5}{c}{en\_disease} \\ 
    & $\text{P}_k$ $\downarrow$ & F1 $\uparrow$ & R-1 $\uparrow$ & R-2 $\uparrow$ & R-L $\uparrow$ &  $\text{P}_k$ $\downarrow$ & F1 $\uparrow$ & R-1 $\uparrow$ & R-2 $\uparrow$ & R-L $\uparrow$\\
    \midrule
    SECTOR         & 14.4 & 71.6 & - & - & - & 26.3 & 55.8  & - & - & - \\
    S-LSTM  & 9.1 & 76.1 & - & - & - & 20.0 & 59.3 & - & - & - \\
    $\text{Transformer}^\text{2}_\text{Bert}$   & 8.2 & - & - & - & - & 18.8  & - & - & - & - \\
    
    \midrule
     
    Naive DS ($\lcls$) & 8.2 & - & - & - & - & 33.5 &  - & - & - & -  \\
    Naive SS ($\lgenseglabel$)  & 9.2 & 73.1 & 79.8 & 62.0 & 79.42 & 24.8 & 38.8 & 52.0 & 30.8 & 51.7  \\
    
    \midrule
    
    Pre-trained DS ($\lcls$) & \bf 6.8  & - & - & - & - & 15.3 & - & - & - & - \\
    Pre-trained SS ($\lgenseglabel$) &	7.1  & \bf	82.3 &	\bf 82.1 & \bf	65.0 &	\bf 81.7 &  \bf 15.0 &	\bf 68.3 &	\bf 62.6 &	\bf 38.5 &	\bf 62.2 \\
    
    \bottomrule
    \end{tabular}
\caption{Wikisection Test set results compared to previous state-of-the-art models. 
SECTOR refers to the best models from \citep{arnold2019sector}, S-LSTM is from \citep{barrow2020joint}, and  $\text{Transformer}^\text{2}_\text{Bert}$ from \citep{lo-2021-transformer-pre}. DS refers to Discriminative Segmentation and SS refers to Structured Summarization. 
}
\label{tab:wikisection}
\end{table*}

\label{sec:wikisection}
\subsection{Mid- to Low-resource, Document Setting}
In order to test our hypothesis that generative segmentation and labeling is competitive for mid- to low- resource settings, we evaluate the English parts of the WikiSection dataset. We set up two types of experiments, namely naive and pretrained. The former uses a Long T5-base as described in Section \ref{sec:expts}, whereas for the latter we pretrain Long T5-base on Wiki-727k with $\lgenseglabel$ and then fine-tune on \emph{en\_city} and \emph{en\_disease}. We used sentence similarity as described in Section \ref{sec:contriever} of the Appendix to verify that there was no data leakage between the Wiki-727k train (pretraining) and Wikisection test (finetuning) datasets.

We evaluate each of these models with discriminative segmentation and structured summarization. Note that in these sets of experiments, we compare discriminative labeling from prior work to our generative labeling. Hence, along with the metrics used in Section \ref{high-res}, we also report the micro-averaged F1 for the task of segment labeling, similar to \citep{barrow2020joint} and \citep{lo2021transformer}.

The results are shown in Table \ref{tab:wikisection}. Finetuning over Wiki-727K  gives us the best structured summarization model for \emph{en\_disease}. For \emph{en\_city}, we notice that pretrained discriminative segmentation does result in a better $\text{P}_k$ than pretrained structured summarization, but only marginally so. For all the labeling metrics, namely micro-averaged F1 and the rouge metrics, both naive and pretrained structured summarization models perform better than the prior work, for \emph{en\_disease} as well as \emph{en\_city}.

\begin{table}[ht!]
\small
	
\setlength{\tabcolsep}{4pt} 
\renewcommand{\arraystretch}{1.3} 
\centering
    \begin{tabular}{l|ccccc}
    \toprule
    \multirow{1}{*}{Model}  &  $\text{P}_k$$\downarrow$  & R-1$\uparrow$ & R-2$\uparrow$ & R-L$\uparrow$ \\
    \midrule
    DialogLM & 38.0  & - & - & - \\

    \midrule
     
    Naive DS ($\lcls$) &   36.7 & - & - & -  \\
    Naive SS ($\lgenseglabel$)   &  44.4 & 19.8 & 6.6 & 17.9 \\
    
    \midrule
    
    Pre-trained DS ($\lcls$) & 33 & - & - & -  \\
    Pre-trained SS ($\lgenseglabel$) & \bf 32.8 & \bf 27.0 & \bf 11.1 & \bf 24.2  \\
    \bottomrule
    \end{tabular}
\caption{QMSUM Test Set segmentation results. DialogLM is from \citep{zhong2021dialoglm}. DS refers to Discriminative Segmentation and SS refers to Structured Summarization. All DS and SS metrics are averaged over 7 runs.
}
\label{tab:qmsum}
\end{table}

\subsection{Low-resource, Conversational Setting}
Finally, we aim to establish the use of structured summarization in a low-resource setting for a different text modality, namely conversational. In this setting, segmentation regards each conversation turn as one "sentence", and predicts positions of turns at segment boundaries. The QMSum dataset is very low-resource, comprising only $157$ training examples.

In the QMSum dataset, we deal with two interrelated problems: presence of very large number of turns (>1000), and large number of input tokens even when we truncate the number of tokens in a turn to a reasonably large number (200). The first problem is detrimental to generative segmentation, since a Wiki-727K-pretrained decoder can learn to map sentence positions up to the number of sentences encountered in Wiki-727K training set. To overcome this, we pretrain on Wiki-727K with a modification: We prepend a random number (up to $1000$) of empty sentences to each training example, thereby increasing learnable sentence positions without affecting other task parameters. To address the second problem, we set a maximum number of tokens ($95$) per turn that keeps all test example inputs under $32768$ tokens, while augmenting the training set with replicas with different maximum number of tokens per turn for each new replica (we use 20, 50, and 200). The rationale is to expose the network to varying degrees of truncation of turn tokens for robustness.

The results are shown in Table \ref{tab:qmsum}. Given very few training sentences, structured summarization performs labeling well as judged by Rouge metrics. For segmentation, although naive structured summarization isn't able to perform better than prior work, when pretrained on Wiki-727K, it sets a new state-of-the-art with generative segmentation, achieving a large performance improvement from the previous best approach. We note that although pretraining was done using a quite distinct text modality (descriptive documents), this facilitates major gains for the conversational modality.

\section{Prior Work}
Our work has most overlap with the text segmentation literature, wherein the task is sometimes called topic segmentation (e.g. \citet{takanobu2018weakly}). The earliest work for segmentation that we're aware of is TextTiling \citep{hearst1994multi}, which draws on the insight that sentences belonging to the same topic/segment should be more coherent than sentences across segments, leading to an unsupervised method for segmentation. Following this work, many other methods that exploit intra-segment coherence/similarity were proposed, including works of \citet{choi2000advances}, \citet{malioutov2006minimum}, \citet{glavavs2016unsupervised}, and more recent works that combine coherence objectives with neural networks, such as \citet{xing2020improving}, \citet{xing2021improving}, \citet{somasundaran2020two}, and \citet{solbiati2021unsupervised}. There are also works for topic segmentation that draw on topic modeling. These works typically learn latent topic representations for segments, such as \citet{riedl2012topictiling}, \citet{misra2009text}.

With the introduction of large enough segmentation datasets for the document modality, such as Wiki-727K \citep{koshorek2018text} and WikiSection \citep{arnold2019sector}, recent work has shifted focus on supervised text segmentation for documents. These works train neural networks with the segment boundary supervision from the datasets. 
Examples from this category include works of \citet{koshorek2018text}, \citet{arnold2019sector}, \citet{xing2020improving}, \citet{barrow2020joint}, \citet{somasundaran2020two}, and \citet{lo2021transformer}.

Whereas document segmentation enjoyed gains facilitated by large datasets, conversation segmentation has been most successfully performed in an unsupervised way to this day. Works in the conversation modality heavily depend on intra-segment coherence, with recent examples being \citet{xing2021improving}, \citet{solbiati2021unsupervised}. \citet{zhong2021dialoglm} perform segmentation on the QMSum dataset with standard supervision, although with limited success, as evidenced by the accuracy gap compared to our method (see Table \ref{tab:qmsum}).

Some works tackle generating labels in the context of segmentation. Most works that used publicly available datasets developed discriminative labeling techniques, such as \citet{arnold2019sector} and \citet{barrow2020joint}. Some works also aimed at solving the labeling task using generation methods. \citet{zhang2019outline} introduced "document outline generation", where they used multiple GRU networks, separately for segmentation and label generation, while using "segment-aware" attention to constrain generation. \citet{liu2021end} use an encoder-decoder transformer for segmenting and labeling news articles. Their method is closely related to our setup, although they use discriminative-only segmentation, and utilize the decoder for only generating labels.

\section{Conclusion}
We introduced a generally applicable technique for unifying text segmentation and segment labeling as a single sequence generation task (\emph{structured summarization}). We have shown that the task is suited to modern-day transformers that handle long inputs, and that one can achieve state-of-the-art performance for both segmentation and labeling across data scales and text modalities. We hope that these results will guide the field towards more general methods that perform structured summarization in research as well as in production settings.

\clearpage
\section{Ethics Statement}
We note that any generative AI technology has the potential to produce harmful, misleading, or offensive content. This should be a guiding principle when considering adopting the technology into real-life use cases.
\bibliography{anthology,custom}
\bibliographystyle{acl_natbib}

\appendix

\begin{large}
\bf {Appendix for "Structured Summarization: Unified Text Segmentation and Segment
Labeling as a Generation Task"}
\end{large}

\section{Encoding sentence positions by reuse of vocab tokens improves over naive models}
\label{sec:sentpos}
As mentioned in Section \ref{sec:decoding}, we validate the use of our simple approach to encode sentence positions in the model. We first note that the native LongT5 tokenizer doesn't use a dedicated BOS token. Therefore, when not using our approach, there is no BOS token we can use for discriminative segmentation. Regardless, we train models to classify the first token of each sentence. Additionally, we train models where we use the EOS token instead for segment boundary classification. We then compare both models to the model where we encode sentence positions according to Section \ref{sec:decoding}. The results are in Table \ref{table:sentpos}. Our approach substantially improves over the naive approach (when using either EOS or BOS), bringing generative segmentation accuracy to the same accuracy of discriminative segmentation within the same model.

\begin{table}[ht]
\small
\centering
\renewcommand{\arraystretch}{1.4} 
\begin{tabular}{l|c}
\toprule
Model & $\text{P}_k$$\downarrow$ \\
 \hline 
naive LongT5 (BOS) & 16.9  \\ 
naive LongT5 (EOS) & 16.2  \\ 
LongT5 (BOS) + sentence positions & 15.0 \\
\bottomrule
\end{tabular}
\caption{Wiki-727K test set results for models trained with $\lcls+\lgenseglabel$}
\label{table:sentpos}
\end{table}

\section{Generative segmentation leads to accurately predicted sentence positions in the output sequence}
\label{sec:sentpos_noise}
Here we follow up on our claim in Section \ref{sec:decoding} that generative segmentation leads to non-erroneous segmentation output when generated in the list of tokens at the NN output. To back this claim, we calculated the fraction of time any output sequence includes an invalid sentence boundary position. An example could be "10 | 31 | 413" in a text with only 300 sentences (last segment boundary is over 300). Another example could be one wherein the output sequence has a non-integer-convertible component, like in "10 | 31e | 299". In Table \ref{table:sentpos_nonnumeric}, we show this erroneous fraction for structured summarization models when tested on Wiki-727K, WikiSection, and QMSum. From the table, it is clear that transformer decoders are easily able to generate tokens that represent integers within the bounds of the task semantics.

\begin{table}[h]
\small
\centering
\renewcommand{\arraystretch}{1.4} 
\begin{tabular}{cccc}
\toprule
Wiki-727K & en\_city & en\_disease & QMSum \\
 \hline 
0.0001 & 0.0025 & 0 & 0 \\ 
\bottomrule
\end{tabular}
\caption{Fraction of examples with at least one erroneous segment boundary position. This is for the structured summarization models, when tested on the respective test set.}
\label{table:sentpos_nonnumeric}
\end{table}

\label{sec:contriever}
\section{Verifying no data leakage between pretraining train- and finetuning test- datasets}

Both Wiki727-K and Wikisection datasets are derived from Wikipedia, and thus pretraining on one and finetuning on the other had a risk of exposing the model to examples it will encounter in the test set.
For part of the Wikisection experiments which use a pretrained model as described in Section \ref{tab:wikisection}, we use such a setup.

To mitigate the above-mentioned risk, we use Contriever \cite{izacard2021contriever} to check for text similarity between the Wiki727-K train examples and Wikisection \emph{en\_disease} and \emph{en\_city} test examples to find examples that need to be eliminated from the Wiki727-K train set. In result, do not find any pairs with exact matches or alarmingly high cosine similarity.

\section{Details on compute and training the models}
We used 8 A100 GPUs (40Gb memory) for training Wiki-727K and WikiSection models, and 1 A100 GPU for training QMSum models. The neural network we use (LongT5) has 220M parameters. 
One epoch of training took about 15 hrs on Wiki-727K, 30 minutes on Wikisection-en\_city, 6 minutes on WikiSection-en\_disease, and 20 minutes on QMSum.
In all experiments, we report test results for the epoch for which the validation metric was best.  The metric of interest for all models was $\text{P}_k$ where applicable, and Rouge otherwise. We performed hyperparameter tuning by selecting runs with the best validation metric. Instead of a full grid search, we used human judgement (over a few 10s of runs) and observed that results were mostly robust to modest changes in most parameters. Our manual approach is mostly necessitated by the lack of sufficiently large computing infrastructure. The only hyperparameters we tuned were learning rate and batch size. We set the maximum number of training epochs to 10 for all experiments.

\end{document}